\documentclass[conference]{IEEEtran}

\usepackage{graphicx,subfigure}
\graphicspath{ {./figures/} }

\ifCLASSINFOpdf
\else
\fi
\begin{document}
%
\title{Hierarchical Clustering in Face Similarity Score Space}

\author{\IEEEauthorblockN{Jason Grant and Patrick Flynn}
\IEEEauthorblockA{Department of Computer Science and Engineering\\
University of Notre Dame\\
Notre Dame, IN 46556}
}


%


\maketitle

\begin{abstract}
Similarity scores in face recognition represent the proximity between pairs
of images as computed by a matching algorithm. Given a large set of images
and the proximities between all pairs, a similarity score space is defined.
Cluster analysis was applied to the similarity score space to develop
various taxonomies. Given the number of subjects in the dataset, we used
hierarchical methods to aggregate images of the same subject.
We also explored the hierarchy above and below the subject level,
including clusters that reflect gender and ethnicity. Evidence supports
the existence of clustering by race, gender, subject, and illumination condition.
\end{abstract}


%
\IEEEpeerreviewmaketitle

\section{Introduction} 

Face recognition is an appealing modality of biometrics because it seeks to mimic the natural way in which
individuals identify one another.  From a recognition standpoint, images of faces can be
acquired easily and non-intrusively, which can be especially beneficial when dealing
with uncooperative subjects. Furthermore, images containing a person's face are
readily available.

Face recognition is not without its challenges, however. 
As the human body ages, the appearance of a person's face also changes. Changes in body weight
affect the shape of one's face, and one's appearance may also be altered by facial hair and 
wrinkles\cite{aging}. Two persons can appear very similar,
as with the case of identical twins, and recognition can be very difficult in non-ideal conditions\cite{twins}.
Recognition under variable lighting and pose also creates challenges
for recognition systems\cite{lightingpose}. 
%

Early deployment of face recognition software was primarily used in security applications such as
identifying persons on a watch list or verifying a person's claimed identity.  Today, commercial
applications of face recognition are becoming just as popular. In mobile applications, face recognition 
is used to secure content on cellular phones. Clustering and organizing of consumer 
photos\cite{leveraging}\cite{consumerphotos} can be seen in applications such as Google Picasa and Facebook.
In these applications, images of faces are clustered at the subject level (images of the same 
person are grouped together). Our research aims to explore the clusters that form above
and below these clustering levels. We hypothesize these clusters to characterize images having the same gender,
ethnicity, illumination, or expression. Furthermore, we attempt to partition the dataset into clusters of
subjects and explore attributes of images which are incorrectly clustered. 
Using the similarity scores of pairs of images from recognition algorithms,
also known as match scores, we utilized hierarchical clustering schemes to explore this ``face space.''

This paper begins with an introduction to hierarchical clustering and related work in face clustering. 
The dataset and clustering methods used are then presented.  Initial analyses of the clustering methods chosen 
are discussed and the experimental setup is described. Afterwards, results from hierarchical
clustering for classification of race, gender, illumination, and expression are presented with analysis.

\section{Background and Related Work}

Hierarchical clustering\cite{dubesjain} is a method of partitioning a set of objects into mutually exclusive groups, containing members
that are homogeneous with respect to some criterion. This technique is considered intrinsic, also known as unsupervised, because 
no category labels denoting \emph{a piori} information are used in a training step. Hierarchical clustering is particularly useful for 
determining an ordering of objects.  Some applications described by Ward include the establishment of taxonomies of plants
and animals with respect to genetic backgrounds, cataloging and organizing materials such as library documents, and in 
identifying job ``types'' and ``subtypes''\cite{ward63}.

Hierarchical clustering may be performed in a top-down or bottom-up fashion. When objects begin in their own cluster and 
are merged into larger clusters as the algorithm works, the method is agglomerative.  A divisive clustering algorithm 
performs these steps in the reverse order, beginning with a single cluster containing all objects and dividing the clusters 
into smaller subclusters.  The results of hierarchical  clustering algorithms are often displayed as dendrograms, 
which are tree structures used to present the arrangement of objects. A cut across the dendrogram creates clusters.

Our goal is to apply these principles to similarity scores for frontal face images. While it is possible
to classify people in many ways, the scope of our classifications is limited to the information available
though a cropped face image. This information includes gender, ethnicity, expression, the presence or absence of eyeglasses, and
the environment in which the image was taken.

A simple procedure for evaluating a clustering method is to compare the category assigned labels of objects within the same cluster.  
Each image within our dataset is assigned a sequence and subject number.  Images with the same subject number belong to the same 
person. Other information such as a subject's year of birth, ethnicity, and gender are stored in a table corresponding to the image sequence
and subject identification number.

Limited research has been conducted in relation to establishing a hierarchy of attributes of face images.
In~\cite{clustersift}, hierarchical face clustering of face images is performed using SIFT features.
Fan and Yeung explore hierarchical clustering as a method to group similar faces for recognition\cite{fan2006}.
Other face clustering methods and applications are explored in~\cite{barr2012}; however, none of this
previous research explores clustering above or below the subject level.

\section{Overview of Hierarchical Clustering Methods}
The following sections will detail the three clustering methods for which we present data: single-link, complete-link, and 
Ward's method. A longer description can be found in~\cite{dubesjain}. These sections will be followed by a synopsis of multidimensional scaling and its use in face image clustering.

Single-link and complete-link clustering are agglomerative methods. Each object begins in 
its own cluster.  This is the first of a three-step process.  The next step is merging, which defines the difference 
between the single-link and complete-link methods.  To begin merging, the lowest threshold value is selected and a 
threshold graph is created. The ordering in which objects are connected in the threshold graph forms the
hierarchy.  The merging height of two objects is equal to the threshold value of 
the threshold graph in which the two objects were joined. 

The complete-link method is more complex. In this method, objects are merged in the hierarchical structure
when a clique is formed among them.  A clique is a fully connected subgraph. The final step is determining 
the stopping criteria. Since a link is formed at each stage of the threshold graph in the single-link method, 
this process takes ($n-1$) iterations, where $n$ is equal to the number of objects. The complete-link method 
terminates when there is an edge between every pair of objects, a total of $n(n-1)/2$ iterations.

Ward's method of clustering differs from that of the single and complete-link approaches by using statistical measures
to compute clusters.  Ward's method is also an agglomerative approach; however, unlike the two previously described
methods that required a threshold graph, this method uses an analysis of variance approach to evaluate the hierarchy and
distance between objects. At each iteration, objects are assigned to clusters in a manner that minimizes the sum of the square 
distances from all objects to their cluster centers.

\section{Dataset}

\begin{figure}
	\centering
        \subfigure[Controlled Neutral]{\label{imgcn}\includegraphics[width=.23\textwidth]{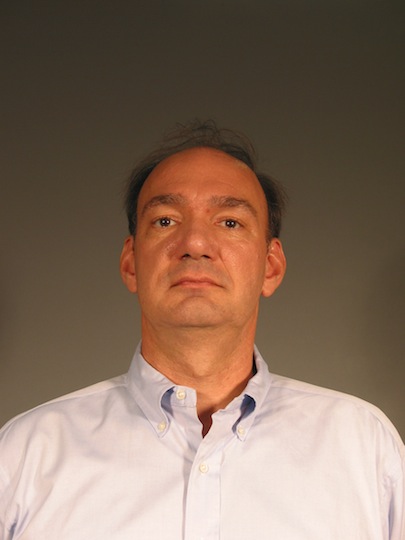} }
        \subfigure[Controlled Smiling]{\label{imgcs}\includegraphics[width=.23\textwidth]{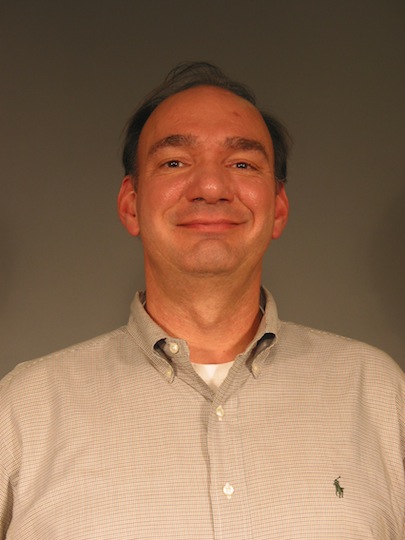} }
        
	\subfigure[Uncontrolled Neutral]{\label{imgun}\includegraphics[width=.23\textwidth]{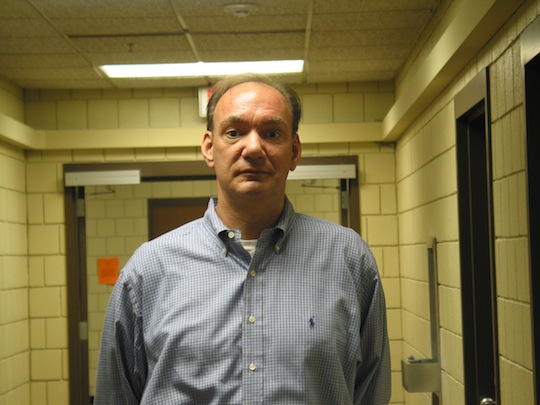} }
        \subfigure[Uncontrolled Smiling]{\label{imgus}\includegraphics[width=.23\textwidth]{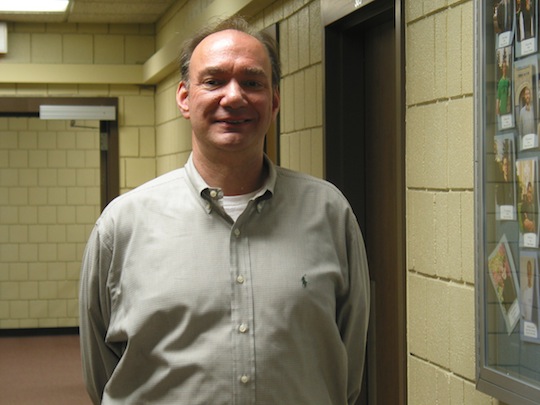} }
	\caption{Sample images of taken from a subject appearing in FRGC ver2.0 dataset. A total of four
		constrained and two unconstrained images were taken in a single session.}
	\label{sampleimages}
\end{figure}

Data for this experiment was taken from the Face Recognition Grand Challenge dataset (FRGC 2.0)\cite{overviewfrgc}.    
Only the validation partition was used in this study.  
The validation set consists of data collected from 4,007 subject sessions, with each subject session containing 
four controlled still images, two uncontrolled still images, and one three-dimensional image. Our data consisted only of 
still images.  Controlled stills were taken under two studio lighting conditions, each setting consisting of two facial 
expressions, happiness and neutral.  The uncontrolled images were taken in various locations and conditions such as 
hallways, atria, or outdoors.  Each pair of uncontrolled stills was also taken with the same two expressions.
Example images are shown in Figure~\ref{sampleimages}. 

Templates of the faces in these images were created using the Cognitec FaceVACS system\cite{cognitec}.  Artifacts 
such as harsh illumination and excessive blurriness caused some enrollment failures. Furthermore, we removed from the 
dataset subjects who appeared in fewer than three acquisition sessions.  In total, our subset of 
the FRGC ver2.0 dataset consisted of 23,355 images taken from 383 subjects. The maximum number of images for
a single subject was 132, the minimum was 18,  and the average was 61.   The majority of images used in 
the study were taken from Asian (6,648) and white (14,853) subjects.  
678 images came from individuals identifying themselves as Hispanic, 414 from Asian-Southern, 
150 from African-American, 96 from Asian Middle-Eastern, and 516 did not carry an ethnicity label.  Gender was nearly 
evenly distributed: 12,392 images were male and 10,963 were female. Subjects wearing glasses appeared in 1,361 
images and the other 21,994 images were taken with subjects not wearing glasses. 11,667 images are of subjects 
depicting a neutral expression or blank stare and the other 11,668 
images contain subjects with a prompted smiling expression.

Using the templates created by Cognitec, an all vs. all matching experiment was conducted. The probe and gallery sets
were identical. For each image pair, a score was generated in the range of 0 to 1, higher scores indicating greater 
similarity. Data was output in matrix form. Analysis was performed in R, an open-source software package for statistical 
computing\cite{CRAN}. Hierarchical clustering methods were implemented using the flashClust algorithm\cite{flashClust}. 
To convert the matching score matrix into a usable format for hierarchical clustering, the similarity matrix was 
transformed into a distance matrix by subtracting each entry in the matrix from the maximum element.

\section{Selecting a Hierarchical Clustering Methods Using Preliminary Analysis}

Because there is no universal best clustering method for all data sets, we initially applied all three hierarchical
methods to our dataset to determine which method(s) would perform well. Our initial criterion of a
method's success was predicated on the appearance of a hierarchical structure within the dendrogram. A secondary
criterion for good performance was based on a cluster purity for a particular feature. Preliminary observations of the
single-link and complete-link clustering methods are presented first, followed by our ultimate decision which was
Ward's clustering method.

The single-link method is a ``greedy'' algorithm.  Clusters are formed by linking unlinked objects in closest proximity.  
Therefore, this technique is not robust to outliers and can create a cascade of errors.  For instance, if an outlier 
joins the cluster, objects that are close in proximity to the outlier will continue to join the cluster and increase the 
number of errors.  Therefore, this method tended to create long cluster strings for our data which generally indicates
poor performance. No natural or representative groupings were obtained.

As opposed to the single-link method, the complete-link method is much more ``cautious''. Clusters are based
upon cliques in the distance matrix.  Because of this property, the first few clusters can be produced quite 
accurately.  Outliers still play an important factor in the resulting data. An object that should be classified
with a group could have a large distance to an object already in the group.  Because of this, the complete-link
method will prevent the cluster from being formed until much later. While smaller clusters are tightly connected,
underlying structures with larger granularities form at much higher levels, thus producing a large proportion 
of very tall branches and no natural separation. 

After initial analysis, it was evident that neither the single-link nor the complete-link methods would perform well with
the data that we were attempting to classify.  One of the most difficult issues stemming from both 
methods was the ability to identify a height level at which the dendrogram could be cut to produce useful clusters. 
These issues are understandable given the large number of objects being clustered; however, this led 
us to explore Ward's Method. Unlike the two previous methods, by visual inspection 
it is apparent that some hierarchy exists within the data. After preliminary results, we decided to continue with 
Ward's method for our subsequent analysis.

\section{Analysis of Dataset using Ward's Clustering Method}

One of the many difficult problems in clustering is determining the number of groups into which the data should be 
clustered\cite{dubeshowmany}.  
These different grouping occur at various heights within the dendrogram. Long branches between merges
often indicate a change in classification, but this may be difficult to identify visually.  Horizontal lines across a dendrogram
also intercept a number of branches.  These values may be used to predict a particular height at which the
dendrogram should be cut to produce clusters. 

We used the same approach in clustering face images; however, we were not certain what hypes of hierarchies would be
present. We began by grouping data into two classes: gender, illumination conditions, and expression.
These classes were chosen from the available metadata of the dataset. Our analysis continued in attempting to cluster
images by race. Results are initially shown with seven classes, one class per known race, and continued as a two-class
problem, due to the largely imbalanced class sizes. Next, a 383-cluster solution is presented, which attempts to
classify the dataset into clusters of subjects.  Finally, 383 individual cluster analyses are performed using only the
genuine images of each subject. In this manner, we attempt to separate illumination conditions and expression.

\subsection{Partitioning by Gender, Eyeglasses, or Expression: A Two-Cluster Solution }

In the coarsest level of granularity, the set of images was divided into two clusters. From our initial analysis, 
we observed that images of the same subjects tended to cluster together.  Therefore, we did not expect 
attributes such as similar expression or the presence of glasses to form large clusters. Our experiments 
verified this expectation. Most notably, the two-cluster solution also did not favor gender, which is a primary way
in which humans classify one another.   

\begin{figure}[tbp]
	\centering
	\subfigure[Clustering of images from Asian subjects]{\label{hcasian}\includegraphics[width=0.48\textwidth]{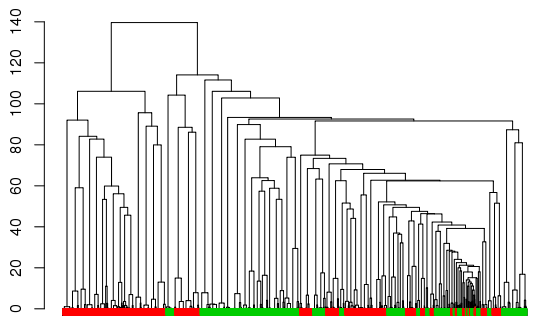}}
	\subfigure[Clustering of images from white subjects]{\label{hcwhite}\includegraphics[width=0.48\textwidth]{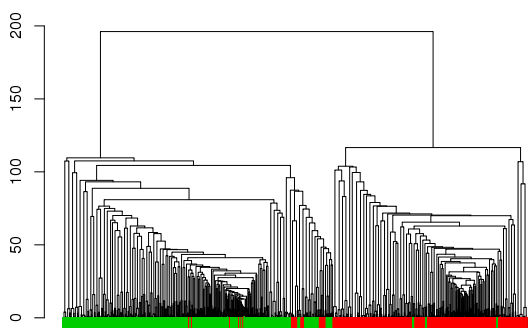}}
	\caption{Shown are the clusterings of Asian and white subjects using Ward's method. Images female subject are shown in
		red and male subjects in green. While there appears to be a 
		natural grouping between white male and female subjects, these observations are not as evident in with
		Asian subjects.}
	\label{hc2groups}
\end{figure}

In a second attempt to classify images by gender, we considered images of a single race at a time.  Images of
Asian and white subjects were clustered separately and then divided into two clusters.  Plots of the two resulting
dendrograms can be seen in Figure~\ref{hc2groups}. Images of white subjects tended to cluster well by gender. The first
cluster consisted primarily of images from  male subjects (7,886 male / 770 female), and the second cluster was consisted
primarily of images from female subjects (6,005 female / 192 male). Images of Asian subjects were not so well behaved.
Grouping the images of Asian subjects into two clusters, the first cluster consisted of 1,476 females and 0 males. However,
the second cluster consisted of 2,046 females and 3,126 males.  Moreover, the structure of the dendrogram (Figure~\ref{hcasian})
presents little evidence to support a secondary divide of the larger, two-gender cluster.

\subsection{Partitioning by Ethnicity: Seven-Cluster and Two-Cluster Solutions}

Noticing that the overall hierarchy using all face images did not favor gender, the data was then divided into seven clusters, 
aiming to bucket each of the reported ethnic groups. Notably, a total of 516 images were taken from 11 subjects that did not 
wish to report an ethnicity. Therefore, while our data is split into seven ethnicities, including one group for unknown ethnicities, 
the number of ethnicities may be in the range of six to seventeen. 

Using a seven-cluster solution, complete separation was not evident; however, some ethnicities tended to cluster together.  
For instance, 3 out of the 7 clusters were comprised purely of Asian subjects and another cluster contained
mostly Asian subjects. Additionally, a cluster contained images of only white subjects and the remaining two clusters were
dominated by images of white subjects. We attempted to incrementally increased the number of clusters from seven to seventeen
in attempt purify clusters; however, this showed no significant increase in separation between the ethnic groups. We attributed this to
the large imbalance of classes.

\begin{figure}[tbp]
	\centering
	\includegraphics[width=0.48\textwidth]{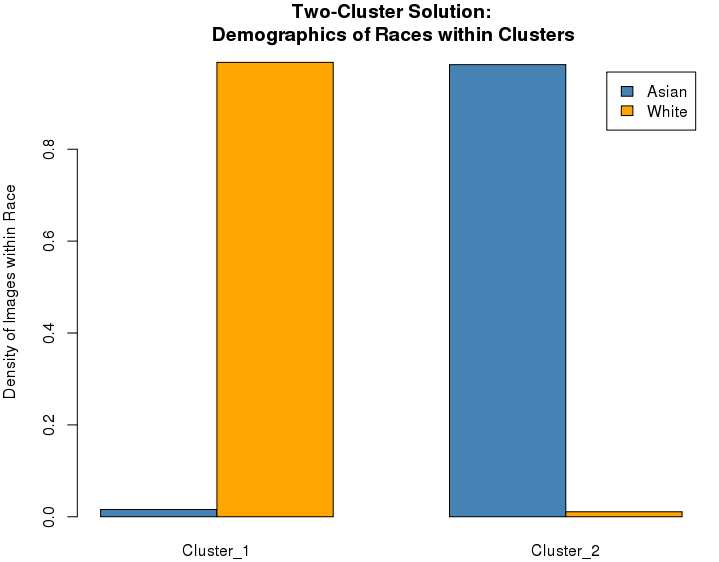}
	\caption{Two-cluster solution using data from only Asian and white subjects.}
	\label{frace2cluster}
\end{figure}

Because our data was dominated by two races and all the clusters of the seven-cluster solution were majority Asian or white subjects,
we provide a two-cluster solution for partitioning by race. As expected, the majority of images of Asian subjects resided in one cluster
and the majority of white subjects resided in the second cluster.
Alternatively, we subsampled the dataset to contain only images of Asian and white subjects and clustered this data into two groups.
Purity of clusters was nearly identical in both demographics (see Figure~\ref{frace2cluster}).

\subsection{Subject-wise Partitioning the Dataset}

Ward's method performed well when separating subjects into clusters of individual subjects. 
Out of 383 clusters, 370 were homogeneous, consisting of one subject (95.8\% purity).  
The set of thirteen error clusters contained 998 images. 788 of those images were of white subjects (5.32\% of white population) and
189 were of Asian subjects (2.84\% of Asian population). 
Surprisingly, illumination conditions did not play a role in misclassification. 
Of the misclassified images, 607 images were taken in controlled illumination and the remaining 259 images were taken
in uncontrolled illumination.  Similarily, misclassified images did not favor gender or expression and the ratio of each class
in the set of misclassified images was approximately equal to the total distribution. 406 images were
of females and 592 were of males, and 503 images were of subjects with a blank stare and 495 were of subjects expressing happiness.

Seven out of 13 clusters containing
multiple subjects were clusters of two subjects, all of which were of the same race.  Six of those seven clusters were of 
subjects of the same gender.  Clusters of multiple races occurred once, in the largest cluster containing 340 images. 
Upon examination, this cluster contained a large number of problematic images that were often out of focus or 
had harsh illumination patterns.

Cluster purity remained nearly the same when considering images of the same race. Using proximity data of just white subjects,
96.1\% of clusters (248/258) consisted of just one subject. Images from Asian subjects fared slightly better; cluster purity for Asian
subjects was 97.8\% (87/89). Notably, the number of white subjects was nearly double that of Asian subjects.

\subsection{Examining the Structure of Proximity Data within Individual Subjects}

Using only the genuine proximity data of a single subject, hierarchical clustering was performed on each subject.  This gave 
us the opportunity to explore the structure of each subject without the possibility of false classifications from other subjects.
Subjects posed in neutral and smiling expressions, and in studio and unconstrained lighting. Using hierarchical clustering we
observed the structure of the proximity data within each subject to see if images of similar illumination conditions or 
expression clustered well.

The dendrograms of subjects whose images were sorted by illumination conditions. These clusterings
show a propensity to group by illumination condition. The same task was performed and images were 
sorted by the subject's expression. A grouping based on expression did not appear evident. This may be explained
by the matching algorithm's robustness to facial expression or by the minimal variation of expression with images.

\section{Conclusions and Future Work}

Through hierarchical methods, we explored underlying structures within similarity scores of face
imagery using the available metadata. When considering a balanced dataset, hierarchical methods proved efficient for
separating the data into two classes of race, white and Asian. We hypothesize that this classification 
could be extended given a larger number of races which were uniformly distributed.
Asian subjects were less likely to cluster by gender when compared to white subjects; however, this was
not attributed to poor matching performance. Comparatively, Asian subjects performed considerably better than
white subjects.
Hierarchical clustering also proved strong in separating groups into classes of subjects.  Given an approximate 
number of subjects within a dataset, hierarchical clustering could be perceived as a potential method for subject grouping 
or cohort analysis\cite{cohort}. 


Improvements to subject clustering may be made using alternative clustering techniques. Our analysis was based on hierarchical methods;
however, other methods include centroid-based clustering, such as k-means\cite{k-means}, or density based clustering\cite{densitybased}.
Clusters formed by by hierarchical methods are created by forming tree-like structures and cutting branches at a constant height value.  
This method is inflexible and may not be well suited for grouping face images.  Studies suggest that individual faces may vary in 
the degree of difficulty for recognition\cite{menagerie}. To account for potential biases of the matching algorithm\cite{otherrace},
our next step will be to cluster fusion scores from multiple face matching algorithms.



%

\bibliographystyle{ieee}
\bibliography{egbib}

\begin{thebibliography}{10}\itemsep=-1pt

\bibitem{cognitec}
{ Cognitec Systems GmbH FaceVACS Software Developers Kit}.
\newblock http://www.cognitec.com.

\bibitem{cohort}
G.~Aggarwal, N.~Ratha, R.~Bolle, and R.~Chellappa.
\newblock Multi-biometric cohort analysis for biometric fusion.
\newblock In {\em IEEE International Conference on Acoustics, Speech and Signal
  Processing}, pages 5224--5227, Apr. 2008.

\bibitem{clustersift}
P.~Antonopoulos, N.~Nikolaidis, and I.~Pitas.
\newblock Hierarchical face clustering using sift image features.
\newblock In {\em IEEE Symposium on Computational Intelligence in Image and
  Signal Processing}, pages 325--329, April 2007.

\bibitem{barr2012}
J.~R. Barr, K.~W. Bowyer, P.~J. Flynn, and S.~Biswas.
\newblock Face recognition from video: A review.
\newblock {\em International Journal of Pattern Recognition and Artificial
  Intelligence}, 26(05), 2012.

\bibitem{dubeshowmany}
R.~C. Dubes.
\newblock How many clusters are best? - an experiment.
\newblock {\em Pattern Recognition}, 20(6):645--663, Nov. 1987.

\bibitem{fan2006}
W.~Fan and D.-Y. Yeung.
\newblock Face recognition with image sets using hierarchically extracted
  exemplars from appearance manifolds.
\newblock In {\em In Proceedings of the 7th International Conference on
  Automatic Face and Gesture Recognition}, pages 177--182, April 2006.

\bibitem{lightingpose}
A.~S. Georghiades, P.~N. Belhumeur, and D.~J. Kriegman.
\newblock From few to many: Illumination cone models for face recognition under
  variable lighting and pose.
\newblock {\em IEEE Transactions on Pattern Analysis and Machine Intelligence},
  23:643--660, 2001.

\bibitem{leveraging}
A.~Girgensohn, J.~Adcock, and L.~Wilcox.
\newblock Leveraging face recognition technology to find and organize photos.
\newblock In {\em Proceedings of the 6th ACM SIGMM International Workshop on
  Multimedia Information Retrieval}, MIR '04, pages 99--106, New York, NY, USA,
  2004. ACM.

\bibitem{consumerphotos}
L.~Gu, T.~Zhang, and X.~Ding.
\newblock Clustering consumer photos based on face recognition.
\newblock In {\em IEEE International Conference on Multimedia and Expo}, pages
  1998--2001, Jul. 2007.

\bibitem{k-means}
J.~A. Hartigan and M.~A. Wong.
\newblock Algorithm as 136: A k-means clustering algorithm.
\newblock {\em Applied statistics}, pages 100--108, 1979.

\bibitem{dubesjain}
A.~K. Jain and R.~C. Dubes.
\newblock {\em Algorithms for clustering data}.
\newblock Prentice-Hall, Inc., Upper Saddle River, NJ, USA, 1988.

\bibitem{densitybased}
H.-P. Kriegel, P.~Kroger, J.~Sander, and A.~Zimek.
\newblock Density-based clustering.
\newblock {\em Wiley Interdisciplinary Reviews: Data Mining and Knowledge
  Discovery}, 1(3):231--240, 2011.

\bibitem{flashClust}
P.~Langfelder and S.~Horvath.
\newblock Fast {R} functions for robust correlations and hierarchical
  clustering.
\newblock {\em Journal of Statistical Software}, 46(11):1--17, 2012.

\bibitem{aging}
H.~Ling, S.~Soatto, N.~Ramanathan, and D.~Jacobs.
\newblock A study of face recognition as people age.
\newblock In {\em IEEE 11th International Conference on Computer Vision}, pages
  1--8, Oct. 2007.

\bibitem{twins}
J.~Paone, P.~Flynn, P.~Phillips, K.~Bowyer, R.~Vorden~Bruegge, P.~Grother,
  G.~Quinn, M.~Pruitt, and J.~Grant.
\newblock Double trouble: Differentiating identical twins by face recognition.
\newblock {\em Information Forensics and Security, IEEE Transactions on},
  9(2):285--295, Feb 2014.

\bibitem{overviewfrgc}
P.~Phillips, K.~Bowyer, T.~Scruggs, E.~Ortiz, J.~Chang, K.~Hoffman, J.~Marques,
  J.~Min, and W.~Worek.
\newblock Overview of the face recognition grand challenge.
\newblock In {\em Computer Vision and Pattern Recognition, 2005. CVPR 2005.
  IEEE Computer Society Conference on}, volume~1, pages 947--954 vol. 1, Jun.
  2005.

\bibitem{otherrace}
P.~J. Phillips, F.~Jiang, A.~Narvekar, J.~Ayyad, and A.~J. O'Toole.
\newblock An other-race effect for face recognition algorithms.
\newblock {\em ACM Transactions Applied Perception}, 8(2):14:1--14:11, Feb.
  2011.

\bibitem{CRAN}
{R Core Team}.
\newblock {\em R: A Language and Environment for Statistical Computing}.
\newblock R Foundation for Statistical Computing, Vienna, Austria, 2012.
\newblock {ISBN} 3-900051-07-0.

\bibitem{ward63}
J.~H. Ward.
\newblock Hierarchical grouping to optimize an objective function.
\newblock {\em Journal of the American Statistical Association},
  58(301):236--244, 1963.

\bibitem{menagerie}
N.~Yager and T.~Dunstone.
\newblock The biometric menagerie.
\newblock {\em IEEE Transactions on Pattern Analysis and Machine Intelligence},
  32(2):220--230, Feb. 2010.

\end{thebibliography}

\end{document}